\documentclass[letterpaper, 10 pt,journal]{IEEEtran}
\usepackage[dvipsnames]{xcolor}
\usepackage{microtype}
\usepackage{amsmath,amsfonts,amssymb}
\usepackage{graphicx}
\usepackage{todonotes}
\usepackage{macros}
\usepackage{balance}
\usepackage{booktabs}
\usepackage{enumitem}
\usepackage{multirow}
\usepackage{subfig}
\usepackage{caption}
\usepackage{mathtools}
\usepackage{cite}
\usepackage{wrapfig}
\usepackage{rotating}

\usetikzlibrary{arrows,patterns,decorations.pathmorphing,positioning,fit,trees,shapes,shadows,automata,calc}

\usepackage[
    pdfauthor={L. Winterer, S. Junges, R. Wimmer, N. Jansen, U. Topcu, B. Becker, J.-P. Katoen},
    pdftitle={Strategy Synthesis in POMDPs via Game-Based Abstractions},
   ]{hyperref}

\newtheorem{theorem}{Theorem}

\begin{document}

\makeatletter
\markboth{Submission to IEEE Transactions on Automatic Control}{}
\makeatother

\title{Strategy Synthesis in POMDPs\\ via Game-Based Abstractions%
\thanks{This work was partly supported by the CDZ project CAP (GZ 1023), by the German Research Foundation (DFG) as part of the
  Cluster of Excellence BrainLinks/BrainTools (EXC\,1086) and as part of the RTG 2236 ``UnRAVeL'', and by the awards ONR \# N000141612051,
  NASA \# NNX17AD04G and DARPA \# W911NF-16-1-0001.}}

\author{Leonore~Winterer, Sebastian~Junges, Ralf~Wimmer, Nils~Jansen, \\ Ufuk~Topcu, Joost-Pieter~Katoen,~\IEEEmembership{Member, IEEE} and Bernd~Becker,~\IEEEmembership{Fellow, IEEE}%
  \thanks{Leonore Winterer and Bernd Becker are with the Albert-Ludwigs-Universit\"at Freiburg, Freiburg im Breisgau, Germany.}%
  \thanks{Ralf Wimmer is with Concept Engineering GmbH and Albert-Ludwigs-Universit\"at Freiburg, Freiburg im Breisgau, Germany.}%
  \thanks{Sebastian Junges and Joost-Pieter Katoen are with the RWTH Aachen University, Aachen, Germany.}%
  \thanks{Nils Jansen is with Radboud University, Nijmegen, The Netherlands.}%
  \thanks{Ufuk Topcu is with The University of Texas at Austin, Austin, TX, USA.}%
  \thanks{\emph{Corresponding author:}\protect\\
    Nils Jansen \protect\\
    Department of Software Science \protect\\
    Radboud University, Nijmegen \protect\\
    Postbus 9010  \protect\\
    6500 GL Nijmegen, The Netherlands \protect\\
    E-mail: \url{n.jansen@science.ru.nl}}%
}
\maketitle


\begin{abstract}
  We study synthesis problems with constraints in
  partially observable Markov decision processes (POMDPs), where the objective is
  to compute a strategy for an agent that is guaranteed to satisfy certain safety and performance specifications.
  Verification and strategy synthesis for POMDPs are, however, computationally intractable in general.
  We alleviate this difficulty by focusing on planning applications and exploiting typical structural properties of such scenarios;
  for instance, we assume that the agent has the ability to observe its own position inside an environment.
  We propose an abstraction refinement framework which turns such a POMDP model into a (fully observable) probabilistic two-player game (PG).
  For the obtained PGs, efficient verification and synthesis tools allow to determine strategies with optimal safety and performance measures,
  which approximate optimal schedulers on the POMDP.
  If the approximation is too coarse to satisfy the given specifications, an refinement scheme improves the computed strategies.
  As a running example, we use planning problems where an agent moves inside an environment with randomly moving obstacles and restricted observability.
  We demonstrate that the proposed method advances the state of the art by solving problems several orders-of-magnitude larger than those that can be handled by existing POMDP solvers.
  Furthermore, this method gives guarantees on safety constraints, which is not supported by the majority of the existing solvers.
\end{abstract}

\begin{IEEEkeywords}
  Robot navigation, POMDP, probabilistic model checking, probabilistic two-player game.
\end{IEEEkeywords}

\section{Introduction}
\label{sec:introduction}
\noindent\IEEEPARstart{P}{artially} observable Markov decision processes (POMDPs) are the formalism of choice to model environments where the current state is not perfectly known~\cite{kaelbling1998planning,thrun2005probabilistic,WongpiromsarnF12}.
They extend Markov decision processes (MDPs), which are non-deterministic models in which an \emph{agent} chooses to perform an action under full knowledge of the environment in which it operates.
The outcome of that action is a probability distribution over the successor states.
In contrast, in a POMDP the agent cannot directly assess the state of the system, but has only access to observations.
By tracking the observations, an agent can infer the likelihood of the environment (and itself) being in a particular state.
This likelihood is called the \emph{belief state} of the agent.
Upon executing an action, the agent updates the belief state according to new observations.
The belief state together with an update function forms a (possibly infinite) MDP, commonly referred to as the underlying \emph{belief MDP}~\cite{ShaniPK13}.
For finite MDPs, tools like \prism~\cite{KNP11} or \tool{Storm}~\cite{DBLP:conf/cav/DehnertJK017} employ efficient model
checking algorithms to assess the probability to reach a certain set of states.
However, due to the potentially infinite belief space, POMDP verification is in general undecidable~\cite{DBLP:journals/ai/MadaniHC03} and also intractable even for rather small instances, and the synthesis of strategies with any given constraints can become a real challenge.

POMDPs are used in a multitude of applications, including control~\cite{WongpiromsarnF12}, scheduling~\cite{NPZ17}, reinforcement learning~\cite{DBLP:journals/corr/Azizzadenesheli16}, and planning~\cite{kaelbling1998planning}.
In this paper, we restrict ourselves to POMDPs describing typical planning scenarios, although it is not unlikely that similar structures may exist in scenarios from many other applications as well.
This restriction allows to exploit certain structural properties, resulting in significantly improved scalability compared to general POMDP
solution approaches. One scenario we are especially interested in is offline planning in dynamically changing environments with uncertainties. 
Here, the goal is to find a \emph{strategy} for an agent that ensures certain desired behavior~\cite{howard1960dynamic}.
As a running example, we take a scenario where a controllable agent needs to traverse a workspace with \emph{obstacles} that can be both \emph{static} or \emph{randomly moving}.
The agent has a limited range of view and can observe moving obstacles only if they are close enough and not hidden behind static obstacles.
A traversal of the workspace is \emph{safe} if the agent avoids any collision.

\paragraph*{Summary of the proposed approach}
\begin{figure*}
\centering
\tikzstyle{block} = [draw, rectangle,
    minimum height=1.0cm, minimum width=1.4cm, align=center]%
    \scriptsize{
\begin{tikzpicture}[node distance=1.9cm]
	\node[block] (prob) {Problem\\description\\ (Sect.~\ref{ssec:problem_statement})};
	\node[block, right=of prob] (pomdp) {POMDP\\(Def.~\ref{def:world_pomdp})};
	\node[block, right=of pomdp] (pg) {PG\\(Def.~\ref{def:abstract_pg})};
	\node[block, right=of pg] (strat) {Optimal strategy\\in PG};
	\node[right=2.8cm of strat] (result) {$[p, \quad p+\tau, \quad \text{optimum},\quad u]$};
	\node[left=0.1cm of result] (txtresult) {induced probability:};
	\draw[->] (prob) -- node[above, align=center] {\textbf{encode}\\cf.~Sect.~\ref{ssec:formal_problem_statement}} (pomdp);
	\draw[->] (pomdp) -- node[above, align=center] {\textbf{abstract}\\cf.~Sect.~\ref{ssec:abstraction}} (pg);
	\draw[->] (pg.60) -- +(0.2, 0.2) -- node[above, align=center] {\textbf{refine} cf.~Sect.~\ref{ssec:refinement}} +(-0.8, 0.2) -- (pg.120);
	\draw[->] (pomdp) -- node[above, align=center] {\textbf{abstract}\\cf.~Sect.~\ref{ssec:abstraction}} (pg);
	\draw[->] (pg) -- node[above,align=center] {\textbf{model checking}\\using \cite{ChenFKPS13}} (strat);
	\draw[dashed] (strat.0) edge[->,bend left=20] node[above] {on PG} (result.172);
	\draw[dashed] (strat.30) edge[->,bend left=20] node[above] {on POMDP} (result.160);
	\draw[dashed,->] (pomdp.270) -- +(0, -0.4) -- node[above, pos=0.75] {assumption: \emph{fully observable}} +(13.3, -0.4) -- (result.351);
\end{tikzpicture}
}
\caption{Schematic overview of the approach}
\label{fig:scheme}
\end{figure*}
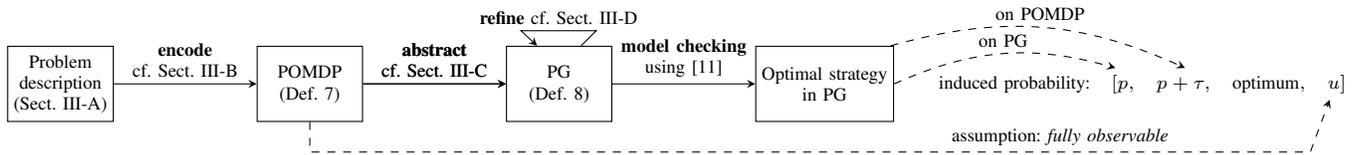
We outline the approach and the structure of the paper in Fig.~\ref{fig:scheme}, and discuss the details in the respective sections. We first \emph{encode} the problem as a POMDP.
Planning scenarios as described above naturally induce certain structural properties in these POMDPs.
In particular, we assume that the agent can observe its own state. On the other hand, the state of the environment, \eg the exact position of the moving obstacles, is observable only if the agent and the obstacle are close according to a given distance metric.
We propose an \emph{abstraction method} that, intuitively, lumps together the states that induce the same observation.
Since it is not exactly known in which state of the environment a certain action is executed, a non-deterministic choice over these lumped states is introduced.
Resolving this choice induces a new level of non-determinism into the system in addition to the choices of the agent:
The POMDP abstraction results in a \emph{probabilistic two-player game} (PG)~\cite{KH09}.
The agent has the role of the first player and chooses an action; the second player determines in which of the possible (concrete) states the action is executed.
When verifying whether a desired behavior is possible in this abstraction, model checking computes, as a byproduct, an \emph{optimal strategy} for the agent on this PG with regard to that behavior.

This \emph{automated} abstraction procedure is inspired by \emph{game-based abstraction}~\cite{KH09,KKNP10,braitling-et-al-vmcai-2015} of potentially infinite MDPs, where states are lumped in a similar fashion.
As quantitative reachability problems are undecidable for POMDPs~\cite{DBLP:journals/ai/MadaniHC03}, our approach is necessarily \emph{incomplete}, as it does not always obtain a strategy that yields the required probability even if one exists.
We show that our approach is \emph{sound}: A strategy for the first player in the PG defines a strategy for the agent in the original POMDP.
Guarantees for the strategy carry over to the POMDP, as, for each strategy, the bounds computed in the PG underapproximate the actual bounds in the POMDP.

We also define a scheme to \emph{refine} the abstraction by considering a history of previous observations. We do this by encoding the last observable position of the moving obstacles into the current state of the game. 
Having access to the last known position limits the possible current positions of the obstacle, mimicking the belief state of the agent. 
Consequently, the second player in the abstraction is more restricted, and thus the method obtains tighter bounds on the POMDP. Therefore, the use of history increases the likelihood of satisfying the given specification. We also modify the environment by placing cameras, effectively increasing the number of observable positions.

We developed a \tool{Python} toolchain, which takes a graph formulation of the workspace as input and implements the proposed abstraction-refinement procedure.
The toolchain uses \prismgames~\cite{ChenFKPS13} as a model checker for PGs.
We created a vast range of examples for the type of planning scenario considered. 
Our preliminary results indicate an improvement by up to three orders of magnitude over the state of the art in POMDP verification~\cite{NPZ17}.

\paragraph*{Contribution}
To summarize our work, we present an abstraction-refinement scheme for POMDPs that iteratively abstracts POMDPs into
probabilistic two-player games. 
We show that our approach is not only sound, but for various examples from our domain, yields --~within minutes of computation time~-- strategies whose quality typically matches those obtained by existing
POMDP solvers.
The method thereby is considerably faster on small models, and scales to significantly larger instances, which could not be analyzed before.

\paragraph*{Related Work}
General verification problems for POMDPs and their decidability are studied
in~\cite{chatterjee2015qualitative,DBLP:journals/ai/MadaniHC03}. A recent survey about decidability
results and algorithms for $\omega$-regular properties is given in \cite{ChatterjeeCT16,ChatterjeeCGK16}.
The probabilistic model checker \prism{} has been extended recently to support POMDPs~\cite{NPZ17}.
Partly based on the methods from~\cite{YB04}, it produces lower and upper bounds for a variety
of queries. Reachability can be analyzed for POMDPs for up to several tens of thousands of states.
An overview on point-based value iteration algorithms for analyzing POMDPs is given in~\cite{ShaniPK13}.
In~\cite{GiroR12}, iterative refinement is proposed to solve POMDPs: Starting with total
information, strategies that depend on unobservable information are excluded.

In~\cite{ZhangWL16}, a compositional framework for reasoning about POMDPs is introduced.
An abstraction-refinement framework based on counterexamples is considered in~\cite{ZhangWL17}. In contrast
to our work, neither \cite{ZhangWL16} nor \cite{ZhangWL17} specializes on planning problems, and there is neither an implementation available
nor any analysis how well these methods scale to systems of relevant size.
Instead of automated abstraction, an interactive human-in-the-loop approach for strategy synthesis in POMDPs is described in~\cite{carr-et-al-ACC-2018},
but such an approach, in contrast to the method described here, may not be fully automated.
The strategies obtained by the method in this paper are finite-memory strategies. 
Finite-memory strategies for POMDPs have been considered in \cite{amato2010optimizing}, but the methods based on non-linear programming may not scale to large state spaces (\eg those on which we demonstrate the proposed method in Sect.~\ref{sec:experiments}).
In contrast, the proposed method utilizes efficient value-iteration-based methods, which are less affected by the size of the state space.
The work in \cite{junges-et-al-uai-2018} studies the relationship between parametric Markov chains (pMCs) and finite-memory strategies for POMDPs;
the close connection between the two formalisms allows to adopt algorithms for pMC synthesis for POMDP strategy computation.
Finally, an overview of applications for PGs is given in~\cite{SvorenovaK16}.

Several methods have been proposed for POMDPs that appear in the context of motion planning.
Sampling-based methods for motion planning in POMDP scenarios are considered
in~\cite{patil2015scaling,burns2007sampling,bry2011rapidly,vasile2016control}.
Other methods employ control techniques to synthesize strategies with safety
considerations under noisy observations and dynamics~\cite{kaelbling1998planning,hauser2010randomized,vitus2011closed}.
Preprocessing of POMDPs in motion planning problems for robotics is suggested
in \cite{GradyMK15}.


This paper is an extended version of~\cite{DBLP:conf/cdc/WintererJW0TK017}, and offers additional
details on our theoretical results as well as extended experiments.

\section{Preliminaries}
\label{sec:foundations}

\subsection{Probabilistic Games}
\label{ssec:prob_games}

\noindent For a finite or countably infinite set $X$, $\dist\colon X\to[0,1]$ denotes a \emph{probability distribution} over $X$ if $\sum_{x\in X}\dist(x) = 1$; the set of all probability distributions over $X$ is $\dists(X)$.
The \emph{Dirac distribution} $\pointdistr{x}\in\dists(X)$ on $x\in X$ is given by $\pointdistr{x}(x) = 1$ and $\pointdistr{x}(y) = 0$ for $y\in X\setminus\{ x \}$.
\begin{definition}[Probabilistic Game]
  \label{def:2pg}
  A \emph{probabilistic game (PG)} is a tuple $\pg = (\states_1, \states_2, \sinit, \acts, \prob)$ where
  $\states = \states_1\dcup\states_2$ is a finite set of \emph{states},
  $\states_1$ the set of states of \emph{Player~1}, $\states_2$ the set of states of \emph{Player~2}, $\sinit\in\states$ the \emph{initial state},
  $\acts$ a finite set of \emph{actions}, and $\prob\colon \states\times\acts\partialto\dists(\states)$ a (partial)
  \emph{probabilistic transition function}.
\end{definition}
Let $\acts(s)=\bigl\{\act\in\acts\,\big|\,(s,\act)\in\dom(\prob)\bigr\}$ denote the \emph{available actions} in $s\in\states$.
We assume that PG $\pg$ is free of deadlock states, \ie $\acts(s)\neq\emptyset$ for all $s\in\states$.
A \emph{Markov decision process} (MDP) $\mdp$ is a PG with $\states_2 = \emptyset$. We write
$\mdp = (\states,\sinit,\acts,\prob)$. A \emph{discrete-time Markov chain} (MC) is an MDP
with $|\acts(s)| = 1$ for all $s\in S$. We write $\dtmc = (\states,\sinit,\prob)$ where $\prob\colon \states\partialto\dists(\states)$.

At Player~1 states (\ie if $s \in\states_1$) Player~1 chooses an available action $\act\in\acts(s)$ non-deterministically, if $s \in\states_2$, Player~2 chooses.
The successor state of $s$ is determined probabilistically according to the probability distribution $\prob(s,\act)$: The probability of $s'$ being the next state is $\prob(s,\act)(s')$.
The state of the game is then $s'$.

A path through $\pg$ is a finite or infinite sequence $\pi = s_0 \xrightarrow{\act_0} s_1 \xrightarrow{\act_1} \cdots$,
where $s_0 = \sinit$, $s_i\in\states$, $\act_i\in\acts(s_i)$, and $\prob(s_i,\act_i)(s_{i+1}) > 0$ for all $i\in\naturals$.
The $(i{+}1)$-th state $s_i$ on $\pi$ is $\pi(i)$, and $\last(\pi)$ denotes
the last state of $\pi$ if $\pi$ is finite. The set of (in)finite paths is $\fpaths[\pg]$ ($\ipaths[\pg]$).

To define a probability measure over the paths of a PG $\pg$, the non-determinism needs to be
resolved by a \emph{strategy} for each player.
\begin{definition}[PG Strategy]
  \label{def:strategy}
  A \emph{strategy} $\sched$ for a PG $\pg$ is a pair $\sched = (\sched_1,\sched_2)$ of functions
  $\sched_i\colon \{\pi\in \fpaths[\pg] \mid\last(\pi)\in S_i\}\to\dists(\acts)$ such that for all $\pi\in \fpaths[\pg]$,
  $\{ \act\mid\sched_i(\pi)(\act)>0\} \subseteq \acts\bigl(\last(\pi)\bigr)$. We also call $\sched_i$ a Player~$i$ strategy. 
  $\scheds[\pg]$ denotes the set of all strategies of $\pg$ and $\scheds[\pg]^i$ all Player~$i$ strategies of $\pg$.
\end{definition}
For MDPs, a strategy consists of a Player~1 strategy only.
A Player~$i$ strategy $\sched_i$ is \emph{memoryless} if $\last(\pi)=\last(\pi')$ implies $\sched_i(\pi)=\sched_i(\pi')$ for all $\pi,\pi'\in\dom(\sched_i)$.
It is \emph{deterministic} if $\sched_i(\pi)$ is a Dirac distribution for all $\pi\in\dom(\sched_i)$.
A \emph{memoryless deterministic} strategy is of the form $\sched_i\colon S_i\to \acts$.

A strategy~$\sched$ for a PG resolves all non-deterministic choices, yielding an \emph{induced MC},
for which a \emph{probability measure} over the set of infinite paths is defined by the standard
cylinder set construction~\cite{BK08}. These notions are analogous for MDPs.

\subsection{Partial Observability}
\label{ssec:partial_obs}

\noindent A partially observable Markov decision processes~\cite{kaelbling1998planning} is obtained by restricting the knowledge of the current state of an MPD.
\begin{definition}[POMDP]
  \label{def:pomdp}
  A \emph{partially observable Markov decision process (POMDP)} is a tuple
  $\pomdp = (\mdp,\obss,\obs)$
  such that
  $\mdp=(\states,\sinit,\acts,\prob)$ is the \emph{underlying MDP of $\pomdp$}, $\obss$ a finite set of
  \emph{observations}, and $\obs\colon \states\to\obss$ the \emph{observation function}.
\end{definition}
W.\,l.\,o.\,g.\ we require that states with the same observations have the same set of
enabled actions, \ie $\obs(s) = \obs(s')$ implies $\acts(s) = \acts(s')$ for all $s,s'\in\states$.
(Otherwise, since the enabled actions are known to the agent, the states could be distinguished, which
contradicts the assumption that the same observations are made in both states.)
More general observation functions $\obs$ have been considered in the literature, taking into account
the last action and providing a distribution over $\obss$. There is a polynomial transformation of
the general case to the POMDP definition used here~\cite{ChatterjeeCGK16}.

The notions of paths and probability measures directly transfer from MDPs to POMDPs.
We lift the observation function to paths: For a POMDP $\pomdp$ and a path
$\pi=s_0\xrightarrow{\act_0} s_1\xrightarrow{\act_1}\cdots s_n\in\fpaths[\pomdp]$, the associated
\emph{observation sequence} is $\obs(\pi)=\obs(s_0)\xrightarrow{\act_0} \obs(s_1)\xrightarrow{\act_1}\cdots\obs(s_n)$.
Note that several paths in the underlying MDP $\mdp$ can give rise to the same observation sequence.
Strategies have to take this restricted observability into account.
\begin{definition}[Observation-Based Strategy]
  \label{def:obsstrategy}
  An \emph{observation-based strategy} of a POMDP $\pomdp$ is a function
  $\sigma\colon \fpaths[\pomdp]\to\dists(\acts)$ such
  that $\sigma$ is a strategy for the underlying MDP and
  for all paths $\pi,\pi'\in\fpaths[\pomdp]$
  with $\obs(\pi) = \obs(\pi')$ we have $\sched(\pi) = \sched(\pi')$.
  $\oscheds$ denotes the set of such strategies.
\end{definition}
An observation-based strategy selects the next action based on the observations and actions made along the current path.

The semantics of a POMDP can be described using a \emph{belief MDP} with an uncountable state space.
Each state of the belief MDP corresponds to a distribution over the states in the POMDP.
This distribution represents the probability of the system to be in any specific state based on the observations made so far.
Initially, the belief is a Dirac distribution on the initial state. In general, a POMDP can be defined as having a set of initial states --
in this case, the initial belief is a uniform distribution over those states.
A formal treatment of belief MDPs is beyond the scope of this paper, for details see \cite{ShaniPK13}.

\subsection{Specifications}
\label{ssec:specifications}

\noindent Given a set $G\subseteq\states$ of \emph{goal states} and a set $B\subseteq\states$
of \emph{bad states}, we consider quantitative reach-avoid properties of the form
$\varphi=\pctlProb_{\geqslant p} (\neg B\ \pctlUntil\ G)$.
This \emph{specification} $\varphi$ is satisfied by a PG if Player~1 has a strategy such that for
all strategies of Player~2 the probability is at least $p$ to reach a goal state without
entering a bad state in between. For POMDPs, $\varphi$ is satisfied if the agent has an
observation-based strategy which leads to a probability of at least $p$ to satisfy $\neg B\ \pctlUntil\ G$.

For MDPs and PGs, memoryless deterministic strategies suffice to prove or disprove
satisfaction of such specifications~\cite{Condon92}. For POMDPs, observation-based strategies
in their full generality are necessary~\cite{Ross83}.

\newcommand{\gacts}{\mathit{Mov}}
\newcommand{\absacts}{\widetilde{\acts}}
\newcommand{\Loc}{\mathit{Loc}}
\newcommand{\loc}{\ensuremath{\ell}}
\newcommand{\visibility}{\ensuremath{\nu}}
\newcommand{\agentstrat}{\ensuremath{\sigma}}
\newcommand{\faraway}{\ensuremath{\mp}}
\newcommand{\Collisions}{\ensuremath{\textsl{Collision}}}
\newcommand{\GoalLocs}{\ensuremath{\textsl{GoalLocations}}}
\newcommand{\Goals}{\ensuremath{\textsl{Goals}}}

\section{Methodology}
\label{sec:methodology}

\noindent We intuitively describe the problem and list the assumptions we make.
After formalizing the setting, we present a formal problem statement.
We explain the intuition behind the concept of game-based abstraction for MDPs, how to apply it to POMDPs, and prove the correctness of our method.

\subsection{The Problem}
\label{ssec:problem_statement}

\noindent We consider a planning problem that involves $n{+}1$ moving agents inside a \emph{world} such as a landscape or a room.
One agent is \emph{controllable} (\agent), the other agents (also called \emph{obstacles}) move probabilistically according to a fixed randomized strategy, which is based on their own location and the location of \agent.
We assume that all agents take turns, moving one after another in a fixed order.
The \emph{position}\footnote{We use the terms \emph{position} and \emph{location} here to avoid confusion with the \emph{states} of the POMDPs and games we use later on.} of an agent defines the \emph{location} inside the world as well as additional properties such as the agent's \emph{orientation}.
A graph captures all possible movements of an agent between positions, referred to as the \emph{world graph} of an agent.
The nodes in the graph uniquely refer to \emph{positions} while multiple nodes may refer to the same \emph{location} in the world.
We require that the graph is free of deadlocks: For every position, there is at least one edge in the graph corresponding to a possible action an agent can execute.

A \emph{collision} occurs, if \agent shares its location with another agent. The set of
\emph{goal nodes} (goal positions) in the graph is uniquely defined by a set of goal
locations in the world. The \emph{target} is to move \agent towards a goal node without
colliding with other agents. Technically, we need to synthesize a strategy for \agent
that maximizes the probability to achieve the target.
Additionally, we make the following assumptions:
\begin{itemize}
  \item The strategies of all obstacles are known to \agent.
  \item \agent is able to observe its own position and knows the goal positions it has to reach.
  \item The positions of obstacles are \emph{observable} for \agent from its current position,
    if they are \emph{visible} with respect to a certain distance metric.
\end{itemize}
Generalizations of the problem statement are discussed in Sect.~\ref{sec:discussion}.

\subsection{Formal Setting}
\label{ssec:formal_problem_statement}

\noindent We first define an individual world graph for each \agent[i] with $0\leq i\leq n$ over a fixed set $\Loc$ of locations.
\begin{definition}[World Graph of {\boldmath\agent[i]}]
  \label{def:world_graph}
  The \emph{world graph} $G_i$ for \agent[i] over
  $\Loc$ is a tuple $G_i = (V_i, v_i^0, \gacts_i,\allowbreak E_i, \loc_i)$
  such that $V_i$ is the set of \emph{positions} and $v_i^0 \in V_i$ is the \emph{initial position} of \agent[i]. $\gacts_i$
  is the set of \emph{movements}\footnote{We use \emph{movements} to avoid confusion with \emph{actions} in PGs.};
  the edges $E_i\colon V_i \times \gacts_i \partialto V_i$ are the \emph{movement effects}.
  The function $\loc_i\colon V_i\to\Loc$ maps a position to the corresponding location.
\end{definition}

\begin{figure}[t]
	\centering
	\subfloat[\scriptsize World graph for Agent~0]{
		\label{fig:WorldGraph:Agent0}
		\scalebox{0.65}{
			\hspace*{1.5cm}
			\includegraphics{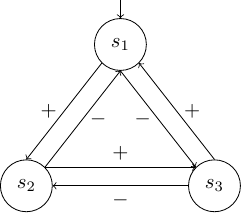}
			\hspace*{1.5cm}
		}
	}
	\subfloat[\scriptsize World graph for Agent~1 under a randomized strategy $\agentstrat_1$]{
		\label{fig:WorldGraph:Agent1}
		\scalebox{0.65}{
			\includegraphics{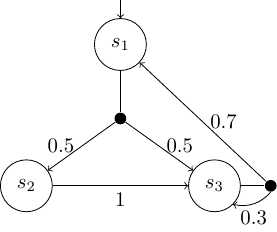}
		}
	}
	\caption{World graphs for two agents and three positions $s_1$, $s_2$ and $s_3$.}
	\label{fig:WorldGraph}
\end{figure}

\noindent The \emph{enabled movements} for Agent~$i$ in position $v$
are $\gacts_i(v) = \bigl\{ m \,\big|\,(v, m)\in\dom(E_i) \bigr\}$.

The viewing range of \agent can be restricted by using a function $\visibility_0\colon V_0\to 2^{\Loc}$ which assigns to each position of \agent
the set of \emph{visible locations}. According to our assumptions, for all $v \in V_0$ it holds that $\loc_0(v) \in \visibility_0(v)$ and $\gacts_i(v) \neq \emptyset$.

Each \agent[i] with $i > 0$ has a randomized strategy $\agentstrat_i\colon V_0 \times V_i \rightarrow \dists(\gacts_i)$,
which maps positions of \agent and \agent[i] to a distribution over enabled movements of \agent[i].

\begin{remark}\label{remark:probabilistic_movements}
It is also possible to allow movements with probabilistic effects in the world graph, \eg to model uncertainty in the behavior of the agents.
As this extension is straightforward, we keep the notations simpler and refrain from allowing probabilistic movements.

\end{remark}

\begin{example}
  Fig.~\ref{fig:WorldGraph} visualizes the concept of world graphs for \agent[0] and one additional agent.
  For the sake of simplicity, we assume in our examples that locations and positions of the agents coincide -- we do not use additional information like the orientation of the agents.
  Both agents move over the same three locations, but their possible movements over these locations are different.
  \agent[0] (Fig.~\ref{fig:WorldGraph:Agent0}) has two enabled movements in each location, so it can move from one location to either of the other two, but not stay in the same location.
  For \agent[1] (Fig.~\ref{fig:WorldGraph:Agent1}), a randomized strategy has already been applied to the world graph.
  Positions $s_1$ and $s_3$, which actually have two enabled movements each, end up with one probabilistic movement instead, while $s_2$ only has one enabled movement to begin with.
  While position $s_2$ has only $s_3$ as a successor, both $s_2$ and $s_3$ can be reached with equal probability from $s_1$; for $s_3$, the probability is higher to move to $s_1$ than to stay in $s_3$.
\end{example}
As \agent has restricted vision, not all parts of the world are observable. 
Therefore, the world graphs for all agents --~with randomized strategies for the obstacles~-- are ultimately subsumed by a single \emph{world POMDP} which has an underlying \emph{world MDP}. 
The MDP models the possible behaviors of all agents based on their associated world graphs.
\begin{definition}[World MDP]
  \label{def:world_mdp}
  For world graphs $G_0,\hdots,G_n$ and strategies $\sched_1,\ldots,\sched_n$,
  the induced \emph{world MDP} $\mdp=(\states,\sinit,\acts,\prob)$ is defined by
  $\states = V_0\times V_1\times\cdots\times V_n\times \{ 0, \ldots, n\}$,
  $\sinit = (v_0^0, v_1^0,\ldots,v_n^0, 0)$, and
  $\acts = \gacts_0 \dcup \{ \bot \}$.
  $\prob$ is defined by:
  \begin{itemize}
  \item For $\act\in\gacts_0(v_0)$ and $\breve{v} \in V_1\times\cdots\times V_n$, we have
      $\prob\bigl((v_0, \breve{v}, 0), \act\bigr) = \pointdistr{(E_0(v_0,\act), \breve{v}, 1)}$.

  \item $\prob\bigl((v_0, \hat{v}, v_i, \breve{v}, i), \bot\bigr)\bigl((v_0, \hat{v}, v'_i, \breve{v}, i{+}1 \mod n{+}1)\bigr) = q$,
    with $\hat{v} \in V_1\times\cdots\times V_{i-1}$,
         $\breve{v} \in V_{i+1}\times\cdots\times V_n$,
         $1 \leq i \leq n$ and
         $q = \sum_{\{ m \,|\, E_i(v_i, m) = v'_i\}} \agentstrat_i(v_0, v_i)(m)$.

  \item $0$ in all other cases.
  \end{itemize}
\end{definition}
\begin{figure}[t]
	\centering
	\scalebox{0.65}{\includegraphics{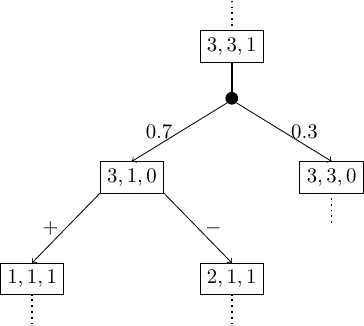}}
	\caption{Cut-out of the world MDP for the world graphs from Fig.~\ref{fig:WorldGraph}}
	\label{fig:WorldMDP}
\end{figure}
The first item in the definition of $\prob$ translates each movement in the world graph of \agent
into an action in the MDP that connects states with probability one, \ie a Dirac distribution is attached to each action.
Upon taking the action, the position of \agent changes and \agent[1] has to move next.

The second item defines movements of the obstacles. In each state where \agent[i] (with $i>0$) is moving next, the
action $\bot$ reflecting this move is added. The outcome of $\bot$ is determined by $\agentstrat_i $ and by Agent $i{+}1$ moving next.

\begin{example}
Fig.~\ref{fig:WorldMDP} shows a cut-out of the world MDP built from the world graphs in Fig.~\ref{fig:WorldGraph:Agent0}
and \ref{fig:WorldGraph:Agent1}. The full world MDP contains 18 different states. The first two numbers
in each state identify the current positions of \agent[0] and \agent[1], respectively; the third entry,
the \emph{turn indicator}, states which agent moves next. \agent[1] moves in a purely probabilistic manner according
to its strategy $\agentstrat_1$, whereas \agent[0] can chose between two different actions, $+$ and $-$. Each agent's movements
only change its own position as well as the turn indicator.
\end{example}

\begin{definition}[World POMDP]
  \label{def:world_pomdp}
  Let $\mdp$ be a world MDP.
  The \emph{world POMDP} $\pomdp = (\mdp,\obss,\obs)$ is defined by $\obss = V_0\times \bigtimes_{1 \leq i \leq n}(V_i \dcup \{ \faraway \})$
  and $\obs$ as follows:
  \[
    \obs((v_0, \hdots, v_n))_i =
      \begin{cases}
         v_i,      & \text{if $\loc(v_i) \in \visibility_0(v_0)$}, \\
         \faraway, & \text{otherwise.}
      \end{cases}
  \]
\end{definition}
The position of \agent[i] is observed if and only if the location of \agent[i] is visible
from the position of \agent.
The reason is that $\loc(v_i)$ describes the location corresponding to the current position of \agent[i], and 
$\visibility_0(v_0)$ is the set of locations visible from the current position of \agent[0].
If the location of \agent[i] is not visible, a value $\faraway$, which is referred
to as \emph{far away}, is observed.

\begin{figure}[t]
    \centering
    \scalebox{0.65}{\includegraphics{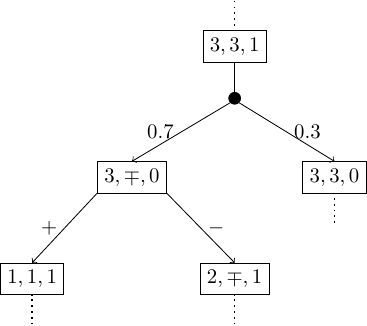}}
    \caption{World POMDP corresponding to Fig.~\ref{fig:WorldMDP}}
    \label{fig:WorldPOMDP}
\end{figure}

\begin{example}
  Fig.~\ref{fig:WorldPOMDP} shows the world POMDP corresponding to the world MDP in Fig.~\ref{fig:WorldMDP}, assuming that
  \agent[0] can only observe the position of \agent[1] when both agents share a location.
  As soon as \agent[1] moves to a different location, it is observed as $\faraway$. For instance, $(3, 1, 0)$ and $(3, 2, 0)$ are
  still distinct states as far as traces in the POMDP are concerned, but as \agent[0] can no longer distinguish between the
  two, it has to behave the same in both states, and we aggregate these states under the same observation label $(3, \faraway, 0)$.
\end{example}
Given a set $\GoalLocs\subseteq\Loc$, the mappings $\loc_i\colon V_i\to\Loc$ are used to define the states corresponding to collisions and goal locations.
In particular, we have $\Collisions =\bigl\{ ((v_0, \hdots, v_n), j)\in S \,\big|\, \exists 1 \leq i \leq n.\, \loc_0(v_0) = \loc_i(v_i) \bigr\}$
and $\Goals = \bigl\{ ((v_0, \hdots, v_n), j) \,\big|\, \loc_0(v_0) \in \GoalLocs \bigr\}$.

\paragraph*{Formal Problem Statement}
We assume that we are given
(1)~a safety threshold $p \in [0, 1]$,
(2)~a world POMDP $\pomdp$ over a set of world graphs $G_0,\ldots,G_n$,
(3)~a set of collision states $\Collisions$, and
(4)~a set of goal states $\Goals$. We define an observation-based strategy $\sched\in\oscheds[\pomdp]$ for $\pomdp$ to be \emph{$p$-safe} for $p\in [0,1]$
if $\pctlProb^{\sched}_{\geqslant p} (\neg \Collisions\ \pctlUntil\ \Goals)$ holds.
The goal is to compute a $p$-safe strategy if one exists.

\subsection{Abstraction}
\label{ssec:abstraction}

\noindent We propose an abstraction method for world POMDPs that builds on \emph{game-based abstraction} (GBAR), originally defined for MDPs~\cite{KH09,KKNP10}.

\paragraph*{GBAR for MDPs}

For an MDP $\mdp = (\states,\sinit,\acts,\prob)$, we assume a partition
$\partition = \{\block_1,\ldots,\block_k\}$ of $S$, \ie a set of non-empty,
pairwise disjoint subsets (called blocks) $\block_i\subseteq\states$ with $\bigcup_{i=1}^k \block_i = \states$.
GBAR takes the partition $\partition$ and turns each block into an abstract state
$\block_i$; these blocks form the Player~1 states. Then $\acts(\block_i) = \bigcup_{s\in\block_i}\acts(s)$.
To determine the outcome of selecting $\act \in \acts(\block_i)$, we add intermediate \emph{selector-states}
$\langle\block_i, \act\rangle$ as Player~2 states. In the selector state $\langle\block_i, \act\rangle$, emanating
actions reflect the choice of the actual state the system is in at $\block_i$, \ie
$\acts\bigl(\langle\block_i, \act\rangle\bigr) = \{s\in\block_i\,|\,\act\in\acts(s)\}$.
For taking an action $s \in B_i$ in $\langle\block_i, \act\rangle$,
the distribution $\prob(s, \act)$ is  lifted to a distribution over abstract states:
\[
  \prob\bigl(\langle\block_i,\act\rangle, s\bigr)\bigl(\block_j\bigr) = \sum_{s' \in \block_j}  P(s, \act)(s').
\]
The semantics of this PG is as follows: For an \emph{abstract state} $\block_i$, Player~1 (controllable)
selects an action to execute. In \emph{selector states},
Player~2 (adversary) selects the worst-case state from $\block_i$ where the selection was executed.

\paragraph*{GBAR for POMDPs}

The key idea in GBAR for POMDPs is to merge states with equal observations.
\begin{definition}[Abstract PG]
  \label{def:abstract_pg}
  The \emph{abstract PG} of POMDP $\pomdp = \bigl((\states,\sinit,\acts,\prob),\obss,\obs\bigr)$ is
  $\pg = (\states_1,\states_2,\sinit',\acts',\prob')$ with
  $\states_1 = \bigl\{ \{ s\in\states\,|\,\obs(s)=\obs(s')\}\,\big|\, s'\in\states\bigr\}$,
  $\states_2 = \bigl\{ \langle\block, \act\rangle\,\big|\,\block\in\states_1\land \act \in \acts(\block) \bigr\}$,
  $\sinit' = \block$ s.\,t.\ $\sinit \in \block$, and
  $\acts' = \states \dcup \acts$.

  The transition probabilities $\prob'$ are defined as follows:
  \begin{itemize}
  \item $\prob'(\block, \act) = \pointdistr{\langle\block,\act\rangle}$ for $\block \in \states_1$ and $\act \in \acts(\block)$,
  \item $\prob'\bigl(\langle\block,\act\rangle, s\bigr)\bigl(B'\bigr) = \sum_{s' \in \block'} \prob(s, \act)(s')$  for $\langle\block,\act\rangle \in \states_2$, $s \in \block$, and $\block' \in \states_1$,
  \item and $0$ in all other cases.
  \end{itemize}
\end{definition}

\begin{figure}[t]
	\centering
	\scalebox{0.65}{\includegraphics{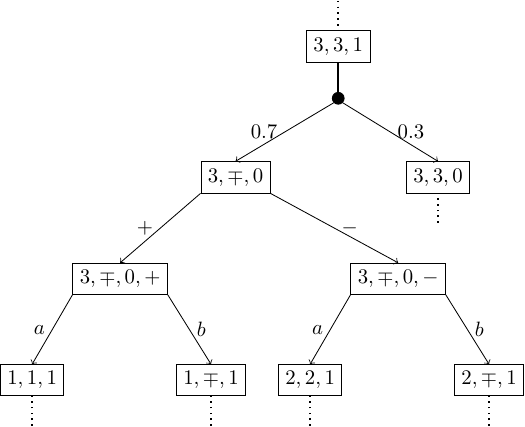}}
	\caption{Abstract world PG corresponding to the world POMDP in Fig.~\ref{fig:WorldPOMDP}.
	Ellipses denote the states of Player~1, rectangles those of Player~2.}
	\label{fig:AbstractPG}
\end{figure}
\noindent By construction, Player~1 has to select the same action for all states in an
abstract state. As the abstract states coincide with the observations, we obtain an observation-based strategy for the POMDP.
For the classes of specifications we consider, a memoryless deterministic strategy suffices for
PGs to achieve the maximal probability of reaching a goal state without collision~\cite{Condon92}.
We thus obtain an optimal strategy $\sched_1\colon S_1\rightarrow\acts$ for Player~1 in the
PG which maps every abstract state to an action. As abstract states are constructed such that
they \emph{coincide with all possible observations in the POMDP} (see Def.~\ref{def:abstract_pg}),
$\sched$ maps every observation to an action.

\paragraph*{Abstract World PG}
We now connect the abstraction to our setting.
For the rest of the section, we assume --~for the ease of representation~-- that there is only one uncontrollable agent,
\ie we have \agent and \agent[1]. Therefore, if \agent sees an agent and moves, no additional
agent will appear. Moreover, \agent either knows the exact state, or does not know where \agent[1] is.

We call the abstract PG of the world POMDP the \emph{abstract world PG}.
The abstract states $B_k$ in the world PG are either of the form $B_k = (v_0, v_1, i)$ or of the form $B_k = (v_0, \faraway, i)$,
with $i\in\{0,1\}$.
In the former, \agent[1] is visible and \agent has full knowledge, in the latter only the own position is known.
Recall that $\faraway$ is a specific value for the distance referred to as \emph{far away}.
Furthermore, all states in an abstract state correspond to the same position of \agent.
For abstract states with full knowledge, there is no non-determinism of Player~2 involved as these states correspond to a single state in the world POMDP.

\begin{example}
  Fig.~\ref{fig:AbstractPG} shows the abstract world PG for our running example from Figs.~\ref{fig:WorldGraph}--\ref{fig:WorldPOMDP}.
  When there is only one action available, we omit the non-deterministic choice.
  In $(3, \faraway, 0)$, Player~1 chooses either $+$ or $-$, leading to
  $(3, \faraway, 0, +)$ or $(3, \faraway, 0, -)$, respectively. Then, Player~2 resolves the \textit{far away} state to the actual
  location $1$ or $2$, choosing the successor state corresponding to the resolved far away state and the action chosen by Player~1.
\end{example}
We extend the notion of \emph{$p$-safety} to strategies in PGs:

\begin{definition}[{\boldmath $p$}-safe strategy]
 A Player~1 strategy $\sched_1\colon S_1\rightarrow\acts$ is \emph{$p$-safe} for $p\in [0,1]$ if $\pctlProb^{(\sched_1,\sched_2)}_{\geqslant p} (\neg \Collisions\ \pctlUntil\ \Goals)$ holds for every Player~2 strategy $\sched_2\colon S_2\rightarrow\acts$.
 \label{def:psafe}
\end{definition}
We formally state the following relationship between strategies in the PG and in the POMDP:
\begin{definition}
  For each memoryless Player~1 strategy $\sigma'$ in the abstract world PG, we define a corresponding observation-based strategy $\sigma$
  in the POMDP by setting $\sigma(o) = \sigma'(\block)$ where $o = \obs(s)$ for all $s \in \block$.
\end{definition}

\begin{proposition}
  \label{prop:correspondence}
  For every path $\pi$ in an abstract world PG under strategy $\sigma'$  with $\pi\models(\neg \Collisions\ \pctlUntil\ \Goals)$, there exists a path $\pi'$ in the world POMDP with $\pi'\models(\neg \Collisions\ \pctlUntil\ \Goals)$.
\end{proposition}

Consider the following intuitive relation between two paths:
Let $\block_0\xrightarrow{\act_0}\langle\block_0,\act_0\rangle\xrightarrow{s\in \block_0}\block_1\xrightarrow{\act_1}\ldots\block_n$  be a path in the PG.
This path is projected to the blocks: $\block_0\xrightarrow{\act_0}\block_1\xrightarrow{\act_1}\ldots\block_n$.
The location of \agent encoded in the blocks is independent of the choices made by Player~2.
The sequence of actions $\act_0\act_1\hdots$ yields a unique path of positions of \agent in its world graph.
Thus, if the path in the PG reaches a goal state, it induces a path in the POMDP which also reaches a goal state.
Choices of Player~2 resolve the non-determinism by selecting a concrete position in place of the abstract one. The actual position \agent[1]
inhabits in the POMDP is always one of the options to pick from. Thus, the abstraction over-approximates the probability for \agent[1] to be in any given location.
Lastly, since \agent can always observe its own location, every collision is observable.
Thus if there is a collision in the POMDP, then there is a collision in the PG.

\begin{figure}[t]
	\centering
	\scalebox{0.65}{\includegraphics{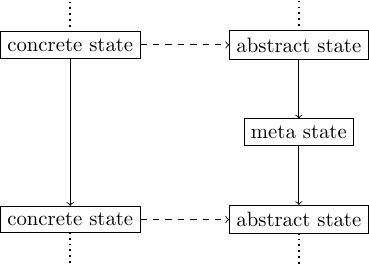}}
	\caption{A graphical representation of the simulation relation.}
	\label{fig:ladder}
\end{figure}

We formalize this relation as a simulation relation as follows:
Let $f\colon\states\rightarrow\states_1$ be an abstraction function that maps each state of the world POMDP to an (abstract) Player~1 state in the PG such that $f(s)=f(s')$ iff $\obs(s)=\obs(s')$ for all $s,s'\in\states$.
Then for each transition $s\rightarrow s'$ in the POMDP, there is a pair of transitions $f(s)\rightarrow s_2 \rightarrow f(s')$, where 
$s_2 \in \states_2$ is a Player~2 state, in the PG. Fig.~\ref{fig:ladder} shows a graphical representation of this relation.

\begin{theorem}
  \label{th:soundness}
  Given a $p$-safe strategy in an abstract world PG, the corresponding strategy in the world POMDP is $p'$-safe
  with $p'\geq p$.
\end{theorem}
\begin{IEEEproof}
Given a $p$-safe Player~1 strategy $\sched_1$ in the abstract world PG, let $\sched$ be the corresponding strategy in the POMDP.
Applying $\sched$ to the POMDP resolves all non-determinism and yields an MC.
Using the simulation relation established before, we can now compute a Player~2 strategy $\sched_2$ for the PG that resolves the non-determinism in a way such that each transition in the MC is mapped to a 2-step-transition in the PG.
The strategy $\sched_1$ is $p$-safe for a given $p\in [0,1]$ if $\pctlProb^{(\sched_1,\sched_2^\ast)}_{\geqslant p} (\neg \Collisions\ \pctlUntil\ \Goals)$ holds for every Player~2 strategy $\sched_2^\ast$ (according to Def.~\ref{def:psafe}).
This property also holds for $\sched_2$ in particular, and as the PG under $\sched_1$ and $\sched_2$ behaves in the same way as the POMDP does under $\sched$, $\sched$ is $p'$-safe with $p'\geq p$.
\end{IEEEproof}

\noindent The assessment of the strategy is conservative:
The abstraction allows behavior of the uncontrollable agents that is not possible in the POMDP.
These so-called \emph{spurious movements} are the result of summing up states with the same abstraction: an agent leaving the visible range in one location could re-appear at (jump to) another location far away, if the non-determinism is resolved in another way.
Thus, $\sched_2$ does not necessarily represent a worst-case strategy in the PG. Therefore the actual probability of safe traversal might be higher in the POMDP than in the PG.

\noindent Applying the strategy to the original POMDP yields a discrete-time Markov chain (MC).
This MC can be efficiently analyzed, \eg by probabilistic model checking to determine the value of $p+\tau$.
Naturally, the optimal strategy for the PG needs not be optimal in the POMDP.

If \agent[1] is visible in a given position, the state of the belief MDP assigns probability $1$ to the corresponding state -- the current state of the POMDP is completely observable, and the successor states in the MDP depend solely on the action choice. These beliefs are represented as single states in the abstract world PG. The abstraction lumps for each position of \agent all (uncountably many) other belief states (in which \agent[1] is not visible) together.

\subsection{Refinement of the PG}
\label{ssec:refinement}

\noindent In the GBAR approach described above, we remove relevant information for an optimal strategy.
As described in the previous section, we strengthen (over-approximate) \agent[1]'s behavior by allowing spurious movements.

If, due to this movements, no safe strategy can be found, the abstraction needs to be \emph{refined}.
In GBAR for MDPs~\cite{KKNP10}, abstract states are split heuristically, yielding a finer over-approximation.
In our construction, we cannot split abstract states arbitrarily: doing so would destroy the one-to-one correspondence between abstract states and observations as seen in Theorem \ref{th:soundness}.
We would thus obtain a partially observable PG, or equivalently, for a strategy in the PG the corresponding strategy in the original POMDP would be no longer observation-based.

However, we can restrict the \emph{spurious movements} of \agent[1] by taking the history of observations made along a path into account. We present three types of \emph{history-based refinements.}

\paragraph{One-Step History Refinement}
If \agent moves to state $s$ from where \agent[1] is no longer visible, we have $\obs(s) = \faraway$.
Upon the next move, \agent[1] could thus appear \emph{anywhere}.
However, until \agent[1] moves, the state of the belief MDP is still Dirac and thus unambiguous; the exact position of \agent[1] is still known, and thereby the positions where \agent[1] can appear.
Similarly, if \agent[1] disappears, the state of the belief MDP is also unambiguous, and upon a turn of \agent in the same direction, \agent[1] will be visible again.

\begin{figure}[t]
	\centering
	\scalebox{0.65}{\includegraphics{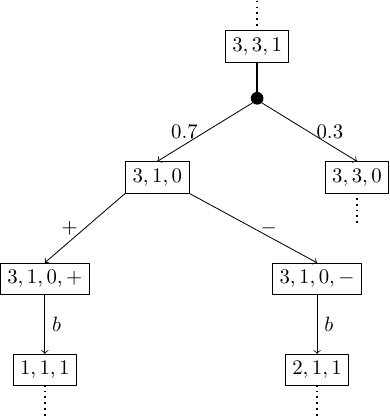}}
	\caption{Abstract world PG with one-step history refinement.
	Ellipses denote the states of Player~1, rectangles those of Player~2.}
	\label{fig:AbstractPGRef}
\end{figure}

The \emph{(one-step history) refined world PG} extends the original PG by additional states $(v_0, v_1, i)$ where $v_1 \not\in \visibility_0(v_0)$, \ie $v_1$ is not visible for \agent.
These ``far away'' states are only reached from states with full information.
Intuitively, although \agent[1] is invisible, its position is remembered for one step.
\begin{example}
  Fig.~\ref{fig:AbstractPGRef} shows the application of one-step history refinement to the Abstract PG from Fig.~\ref{fig:AbstractPG}. When \agent[1] leaves location~3 and is no longer
  visible, it can only be in location~1. Until \agent[1] moves, its location is known to \agent[0].
\end{example}

\paragraph{Multi-Step History Refinement}
Further refinement is possible by considering  longer paths.
If we first observe \agent[1] at location $x$, then loose visibility for one turn, and then observe \agent[1] again at position $y$, then we know that either $x$ and $y$ are at most two moves apart or that such a movement is \emph{spurious}.
To encode the observational history into the states of the abstraction, we store the \emph{last known position} of \agent[1], as well as the \emph{number $m$ of moves} made since then.
We then only allow \agent[1] to appear in positions which are at most $m$ moves away from the last known position. We can cap $m$ by the diameter of the graph.

\paragraph{Region-Based Multi-Step History Refinement}
As the refinement above blows up the state space drastically, we utilize \emph{magnifying lens abstraction}~\cite{AlfaroR07}.
Instead of single locations, we define \emph{regions of locations} together with the information if \agent[1] could be present.
After each move, we extend the possible regions by all neighbor regions.

More formally, the \emph{(multi-step history) refined world PG} has a refined far-away value $\faraway$: Given a partition of the positions of \agent[1], \eg extracted from the graph, into sets $\mathcal{X} = \{ X_1, \hdots, X_l \}$ with $\bigcup_{X \in \mathcal{X}} = V_1$ and  $X_i \cap X_j = \emptyset$ for all $1 \leq i < j \leq l$.
We define  $\faraway'\colon \mathcal{X} \rightarrow \{0,1\}$.
Abstract states now are either of the form $(v_0, v_1, i)$ as before, or $(v_0, \faraway', i)$.
For singleton regions, this refinement approach coincides with the method above.
The region-based approach also offers some flexibility:
If for instance two regions are connected only by the visible area, \agent can assure whether \agent[1] enters the other region.

\begin{theorem}
  \label{th:soundness_refined}
  A $p$-safe strategy in a refined abstract world PG has a corresponding $p$-safe strategy in the world POMDP.
\end{theorem}

\begin{IEEEproof}
First, a deterministic memoryless strategy $\sched'$ on a refined abstract world PG needs to be translated to a strategy $\sched$ for the original POMDP such that $p$-safety is conserved.
Again, we can show a projection of a path $\block_0\xrightarrow{\act_0}\langle\block_0,\act_0\rangle\xrightarrow{s\in \block_0}\block_1\xrightarrow{\act_1}\ldots\block_n$ in the PG to the blocks
$\block_0\xrightarrow{\act_0}\block_1\xrightarrow{\act_1}\ldots\block_n$, and again, the location of \agent[0] is independent of the choices Player~2 makes. However, each block on the path no longer corresponds
solely to the location and \textit{current} observation of \agent[0], but instead to a history of observations. Therefore, the strategy $\sched$ is not memoryless anymore but has a finite memory at most $m$
according to the maximum number of moves that are observed.
Formally, the simulation relation introduced for the unrefined abstraction does no longer map from the abstraction to the POMDP, but from the refined abstraction to a (one- or more-step) history unrolling of the POMDP. All further arguments remain the same.
\end{IEEEproof}
\begin{theorem}
  If an abstract world PG has a $p$-safe strategy, its refined abstract world PG has a $p'$-safe strategy with $p' \geq p$.
\end{theorem}
\begin{IEEEproof}
The proposed refinements eliminate spurious movements of \agent[1] from the original abstract world PG.
Intuitively, the number of states where Player~2 may select states
with belief zero (in the underlying belief MDP) is reduced.
We thus only prevent paths that have probability zero in the POMDP.
Vice versa, the refinement does not restrict the movement of \agent and any path leading to a goal state still leads to one in the refinement.
However, the behavior of \agent[1] is restricted, therefore, the probability of a collision drops.
Intuitively, for the refined PG strategies can be computed that are at least as good as for the original PG.
\end{IEEEproof}

\paragraph{Refinement of the Graph}

\noindent The proposed approach cannot solve every scenario --~the problem is undecidable~\cite{ChatterjeeCT16}.
Therefore, if the method fails to find a $p$-safe strategy, there may still exist such a strategy.
However, we know that if we modify the graph by increasing the view range of the agent (\eg by adding \emph{cameras} to potential blind spots), the maximal level of safety does not decrease in both the POMDP and the PG.

\section{Case Study and Implementation}
\label{sec:case_study}

\newcommand{\robot}{\textsl{R}}
\newcommand{\cleaner}{\textsl{VC}}

\subsection{Description}
\label{ssec:description}

\noindent For our experiments we choose the following scenario:
A (controllable) robot $\robot$ and a vacuum cleaner $\cleaner$ are moving around in a
two-dimensional grid world with static opaque obstacles.
Neither $\robot$ nor $\cleaner$ may leave the grid or visit grid cells occupied by
a static obstacle. Thus, locations correspond to grid cells.
The position of $\robot$ consists of the cell $C_\robot$ and a wind
direction. $\robot$ can move one grid cell forward
or turn by $90^\circ$ in either direction without changing its location.
The position of $\cleaner$ is determined solely by its cell $C_\cleaner$.
In each step, $\cleaner$ can move one cell in any wind direction.
We assume that $\cleaner$ moves to all available successor
cells with equal probability.

\begin{figure}[t]
	\centering
	\subfloat[\scriptsize World graph for $\robot$ in the grid]{
		\label{fig:WorldGraph:R}
		\scalebox{0.4}{
			\hspace*{0.5cm}
			\includegraphics{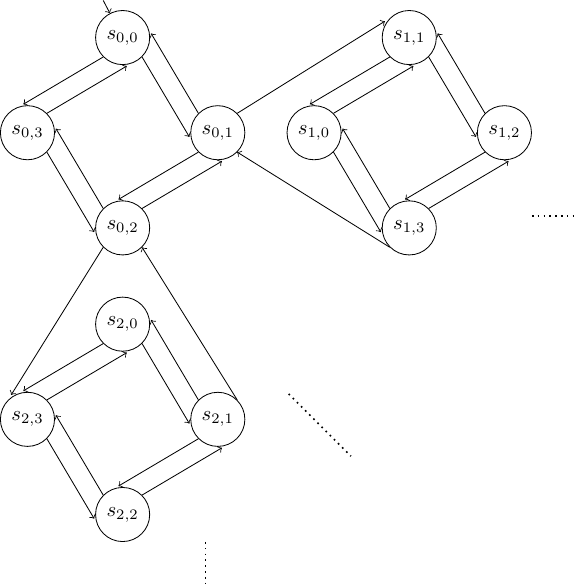}
			\hspace*{0.5cm}
		}
	}
	\subfloat[\scriptsize World graph for $\cleaner$ in the grid]{
		\label{fig:WorldGraph:VC}
		\raisebox{0.5cm}{\scalebox{0.5}{
                        \hspace*{0.5cm}
			\includegraphics{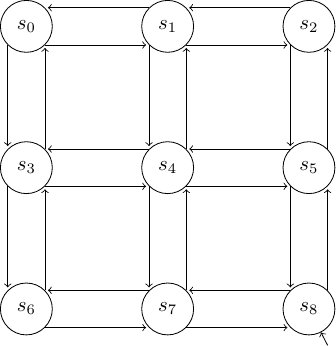}
                        \hspace*{0.5cm}}
		}
	}
	\caption{World graphs for a $3\times 3$ grid as described in the case study.}
	\label{fig:WorldGraphGrid}
\end{figure}

Fig.~\ref{fig:WorldGraphGrid} shows the world graphs describing this scenario for a $3\times 3$ grid.
The $\cleaner$ (Fig.~\ref{fig:WorldGraph:VC}) can move from each cell to each of the adjacent cells.
$\robot$ (Fig.~\ref{fig:WorldGraph:R}) can only move in one direction, depending on its orientation -- each of
the cardinal directions represents one orientation of $\robot$, so four positions of the world graph represent 
the same location (\eg $s_{0,0}$, $s_{0,1}$, $s_{0,2}$ and $s_{0,3}$).
The orientation can be changed by turning $90^\circ$ in either direction.
In particular, if $\robot$ wants to move back to a grid cell it just left (\eg from  $s_{2,3}$ 
back to $s_{0,3}$), it has to turn twice first.

The sensors on $\robot$ only sense $\cleaner$ within a \emph{viewing range} $r$ around $C_\robot$.
More precisely, $\cleaner$ is visible iff $\|C_\robot - C_{\cleaner}\|_{\infty}\leq r$ and
there is no grid cell with a static obstacle on the straight line from $C_\robot$'s center to $C_{\cleaner}$'s center.
That means, $\robot$ can observe the position of the $\cleaner$ if $\cleaner$ is in the viewing range and $\cleaner$
is not hidden behind an obstacle. A refinement of the world can be realized by adding additional cameras, which make
cells visible independent of the location of $\robot$.

\subsection{Toolchain}
\label{ssec:tool_chain}

\noindent To synthesize strategies for the scenario described above, we implemented a \tool{Python} toolchain that has as input a grid with the locations of all obstacles, the location of cameras, and the viewing range.
As output, two \prism{} files are created: A PG formulation of the abstraction including one-step history
refinement, to be analyzed using \prismgames{}~\cite{ChenFKPS13}, and the original POMDP for \prismpomdp{}~\cite{NPZ17}.
For multi-step history refinement, additional regions can be defined.

The encoding of the PG contains a precomputed lookup-table for the visibility relation. The PG is described by two parallel
processes running interleaved: One for Player~1 and one for Player~2. Because only $\robot$ can make choices, they constitute
Player~1's actions; $\cleaner$'s moves form Player~2's actions. More precisely, the process for $\robot$
contains its position, and the process for $\cleaner$ either contains its position or a \emph{far-away value}.
Then, Player~1 makes its decision, afterwards the outcome of the move \emph{and} the outcome of the subsequent move of
$\cleaner$ are compressed into one step of Player~2.

Additionally, to allow a comparison of our results to state-of-the-art point-based POMDP solvers~\cite{ShaniPK13},
the toolchain generates a POMDP in \solvepomdp's~\cite{DBLP:conf/aaai/WalravenS17} input format. Most other point-based solvers (like pomdp-solve\footnote{\url{http://www.pomdp.org/code/index.html}} or DESPOT\footnote{\url{https://github.com/AdaCompNUS/despot}}) only support infinite-horizon discounted maximum reward instead of reach-avoid properties.
In \solvepomdp, however, we can use a discount factor of $\beta = 1$ to compute \emph{undiscounted} maximum reward and use the following construction:
We make all \textit{Collision} and \textit{Target} states absorbing by replacing their outgoing edges with self-loops with probability 1 and reward 0. 
The incoming edges of \textit{Target} states obtain a reward of 1, all other transitions have zero reward.
With undiscounted rewards, $\solvepomdp$ returns the maximum probability of reach-avoid properties.


\section{Experiments}
\label{sec:experiments}

\subsection{Experimental Setup}
\label{subsec:exp_setup}

\begin{figure}
  \centering
  \rotatebox{90}{\includegraphics{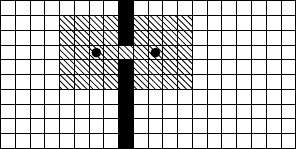}}
  \caption{Grid for \ref{sc:tworooms}. The cameras observe the shaded area.}
  \label{fig:grids}
  \vspace{-4mm}
\end{figure}

\noindent All experiments were run on a machine with a 3.6~GHz
Intel\textsuperscript{\textregistered} Core\textsuperscript{TM} i7-4790 CPU
and 16~GB RAM, running Ubuntu Linux 16.04. We denote experiments taking over 5400~s
CPU time as time-out and taking over 10~GB memory as mem-out (MO).
We considered several variants of the scenario described in Sect.~\ref{ssec:description}.
The robot $\robot$ always starts in the upper-left corner and has the lower-right corner
as target. The $\cleaner$ starts in the lower-right corner. In all variants,
the view range is $3$ and the average number of outgoing transitions per state is about 2.
We evaluated the following five scenarios:
\begin{enumerate}[label={\textbf{SC\arabic*}}, wide, labelwidth=!, labelindent=0pt,topsep=0pt]
\item\label{sc:grid} Rooms of varying size without obstacles.

\item\label{sc:cross} Differently sized workspaces with a cross-shaped obstacle in the center,
  which scales with increasing grid size.

\item\label{sc:random} A $25\times 25$ workspace with up to $70$ randomly placed obstacles.

\item\label{sc:tworooms} A workspace consisting of two rooms (together $10\times 20$) as depicted in Fig.~\ref{fig:grids}.
  The doorway connecting the two rooms is a potential point of failure, as $\robot$
  cannot see to the other side. To improve reachability, we added two cameras for better visibility.

\item\label{sc:corridor} Corridors of the format $4\times x$ -- long, narrow grids the robot has
  to traverse from top to bottom, passing the $\cleaner$.
\end{enumerate}

\setlength\tabcolsep{3pt}
\begin{table*}
 \centering
 \caption{\normalfont Comparing the POMDP solution (\prismpomdp{}) with the PG abstraction solution (\prismgames{}) on \ref{sc:grid}.}
 \label{tab:pomdp_vs_pg}
 \begin{tabular}{crrrrrrrrrrrrrr}
 \toprule
           & \multicolumn{5}{c}{POMDP solution} & \multicolumn{5}{c}{PG solution} & \multicolumn{1}{c}{Lifting} & \multicolumn{1}{c}{MDP}\\
 Grid size & States & Choices  & Result & Model Time & Sol.\ Time & States & Choices  & Result & Model Time & Sol.\ Time & Result & Result \\
 \midrule
 $3\times 3$   &   299  &   515  & \textbf{0.8323} & 0.063 &   \textbf{0.26}    &     396 &     639 & \textbf{0.8323} &   0.075 & \textbf{0.040}    & 0.8323   & 0.8323\\
 $4\times 4$   &   983  &  1778  & \textbf{0.9556} & 0.099 &  \textbf{1.81}     &    1344 &    2192 & \textbf{0.9556} &   0.098 & \textbf{0.078}    & 0.9556   & 0.9556\\
 $5\times 5$   &  2835  &  5207  & \textbf{0.9882} & 0.144 &  \textbf{175.94}   &    6016 &   10448 & \textbf{0.9740} &   0.193 & \textbf{0.452}    & 0.9825   & 0.9882\\
 $5\times 6$   &  4390  &  8126  & \textbf{0.9945} & 0.228 & \textbf{4215.06}   &    7986 &   14199 & \textbf{0.9785} &   0.220 & \textbf{0.534}    & 0.9893   & 0.9945\\
 $6\times 6$   &  6705  & 20086  & \unknown        & 0.377 & \textbf{--\,MO\,--}  &   10544 &   19150 & \textbf{0.9830} &   0.267 & \textbf{1.414}    & 0.9933   & 0.9970\\
 $8\times 8$   & 24893  & 47413  & \unknown        & 1.735 & \textbf{--\,MO\,--}  &   23128 &   43790 & \textbf{0.9897} &   0.470 & \textbf{6.349}    & 0.9992   & 0.9998\\
 $10\times 10$ & 66297  & 127829  & \unknown        & 9.086 & \textbf{--\,MO\,--}  &   40464 &   78054 & \textbf{0.9914} &   0.921 & \textbf{12.652}   & 0.9999   & 0.9999\\
 $20\times 20$ & \multicolumn{5}{c}{-- Time out during model construction --}              &  199144 &  395774 & \textbf{0.9921} &   9.498 & \textbf{127.356}  & 0.9999   & 0.9999\\
 $30\times 30$ & \multicolumn{5}{c}{-- Time out during model construction --}              &  477824 & 957494  & \textbf{0.9921} &  40.929 & \textbf{489.369}  & --\,MO\,-- & 0.9999\\
 $40\times 40$ & \multicolumn{5}{c}{-- Time out during model construction --}              &  876504 & 1763214 & \textbf{0.9921} & 135.551 & \textbf{1726.489} & --\,MO\,-- & 0.9999\\
 $50\times 50$ & \multicolumn{5}{c}{-- Time out during model construction --}              & 1395184 & 2812934 & \textbf{0.9921} & 355.732 & \textbf{3963.281} & --\,MO\,-- & --\,MO\,-- \\
 \bottomrule
 \end{tabular}
 \vspace{-2mm}
\end{table*}

\subsection{Results}
\label{ssec:results}

\noindent Table~\ref{tab:pomdp_vs_pg} shows the direct comparison between the POMDP
description and the abstraction for~\ref{sc:grid}. The first column gives the grid size.
Then, first for the POMDP and afterwards for the PG, the table lists the \emph{number
of states}, \emph{non-deterministic choices}, and \emph{transitions} of the model.
The results include the safety probability induced by the optimal strategy (``\emph{Result}''),
the run times (all in seconds) \prism{} takes for \emph{constructing the state space} from
the symbolic description (``\emph{Model Time}''), and finally the time to solve the POMDP\,/\,PG
(``\emph{Sol. Time}'').

The probability obtained on the PG is a rather pessimistic lower bound, as the abstraction
gives the uncontrollable agent more power than it actually has in the POMDP. 
If we apply the Player~1 strategy that we compute by solving the PG on the POMDP, we obtain better results;
by analyzing its induced MC we obtain improved safety probabilities (``\emph{Lifting Result}``). 
An upper bound on the probability is computed on the fully observable MDP (``\emph{MDP  Result}``). 
An entry ``\emph{--\,MO\,--}'' in either the ``Sol. Time'' or ``Result'' column of one of the solution approaches means that
this particular approach ran out of memory before hitting the time limit. In these cases, the result is unknown.
Note that optimal strategies from this MDP are in general not observation-based and therefore not admissible for the POMDP.
The time for creating the \prism{} files was $< 0.1$~s in all cases.

\begin{figure}[t]
    \centering
    \scalebox{0.65}{\includegraphics{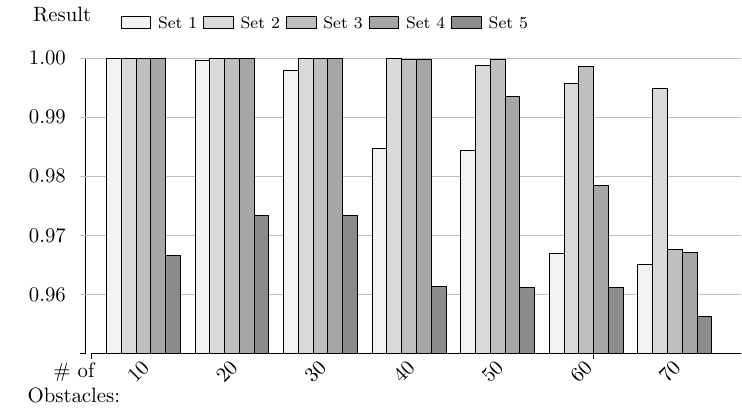}}
    \caption{Results for \ref{sc:random} for different sets of random obstacles}
    \label{fig:pg_random_1}
\end{figure}

\begin{figure}[t]
    \centering
    \scalebox{0.65}{\includegraphics{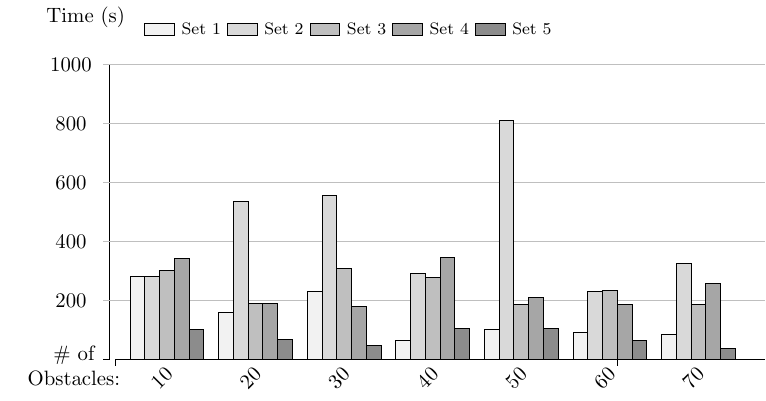}}
    \caption{Solving times for \ref{sc:random} for different sets of random obstacles}
    \label{fig:pg_random_2}
\end{figure}

\begin{figure}[t]
    \centering
    \scalebox{0.65}{\includegraphics{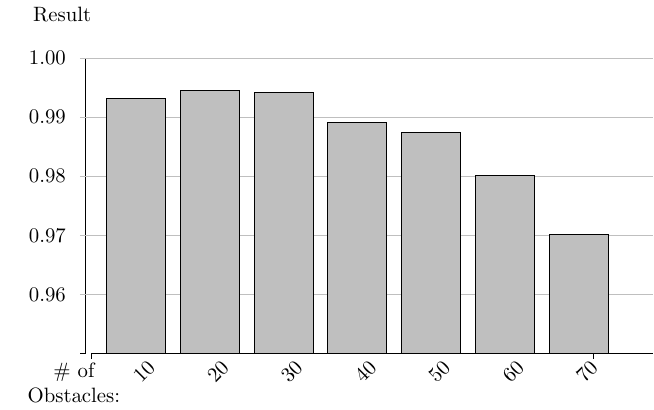}}
    \caption{Accumulated results for \ref{sc:random} for different sets of random obstacles}
    \label{fig:pg_random_3}
\end{figure}

\begin{figure}[t]
    \centering
    \scalebox{0.65}{\includegraphics{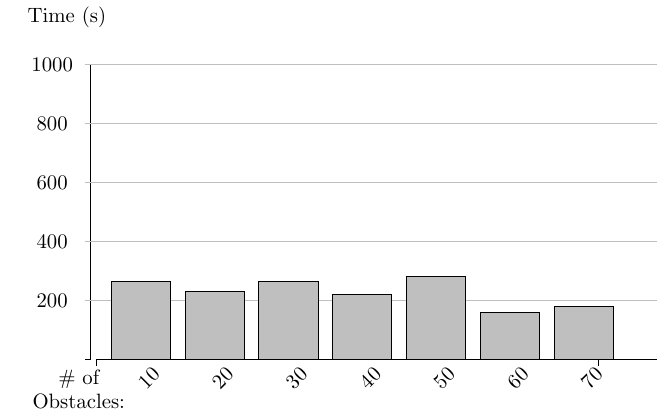}}
    \caption{Accumulated solve times for \ref{sc:random} for different sets of random obstacles}
    \label{fig:pg_random_4}
\end{figure}

Table~\ref{tab:pg_corridor_cross} lists data for the PG constructed from \ref{sc:cross}
(first block of rows) and \ref{sc:corridor} (without additional refinement) in
the second block, analogous to Table~\ref{tab:pomdp_vs_pg}. The third block also lists data for \ref{sc:corridor}, with the addition of 
a simple region-based refinement considering only one step of history (see Sect.~\ref{ssec:refinement}). Additionally the runtime for
creating the symbolic description is given (``Run times\,/\,Create''). On the fully observable MDP, the resulting probability is 1.0 for all \ref{sc:cross}
and 0.999 for all \ref{sc:corridor} instances.
Figures~\ref{fig:pg_random_1}, \ref{fig:pg_random_2}, \ref{fig:pg_random_3} and \ref{fig:pg_random_4} show the results for \ref{sc:random} for
five different sets of random obstacles. 
The data for \ref{sc:tworooms} is shown in Table~\ref{tab:pg_rooms}. Its
structure is identical to that of Table~\ref{tab:pg_corridor_cross},
with the first column (``\#C'') corresponding to the number of cameras added for the graph refinement as described in Sect.~\ref{ssec:refinement}.

\newcommand*\rot[1]{\rotatebox{90}{#1}}
\begin{table}
 \centering
 \caption{\normalfont Results for the PG for differently sized models.}
   \label{tab:pg_corridor_cross}
   \scalebox{0.98}{
   \begin{tabular}{lcrrrrrr}
    \toprule
    &  &  \multicolumn{3}{c}{PG} & \multicolumn{3}{c}{Run times}\\
    & Grid & States & Choices  & Result & Create & Model & Solve\\
    \midrule
    \multirow{4}{*}{\rot{\ref{sc:cross}}} & $11\times 11$ &  36000 &   66978 & \textbf{0.9905} &  0.08 & 3.6 & \textbf{30.7}\\
    & $21\times 21$ & 173400 &  331774 & \textbf{0.9974} &  1.19 &   48.7 & \textbf{158.9}\\
    & $31\times 31$ & 430760 &  834814 & \textbf{0.9978} &  7.62 &  316.9 & \textbf{329.8}\\
    & $41\times 41$ & 808120 & 1576254 & \textbf{0.9978} & 31.92 & 1686.0 & \textbf{1708.9}\\
    \midrule
  \multirow{4}{*}{\rot{\ref{sc:corridor}}}
    & $4\times 40$  &  50880 &  93734 & \textbf{0.9228} & 0.01 & 1.6 & \textbf{37}\\
    & $4\times 60$  &  77560 & 143254 & \textbf{0.8923} & 0.01 & 3.1 & \textbf{41}\\
    & $4\times 80$  & 104240 & 192774 & \textbf{0.8628} & 0.01 & 5.4 & \textbf{128}\\
    & $4\times 100$ & 130920 & 242294 & \textbf{0.8343} & 0.02 & 8.6 & \textbf{101}\\
    \midrule
    \multirow{4}{*}{\rot{\ref{sc:corridor} + ref.}}
    & $4\times 40$  &  68316 &  131858 & \textbf{0.9733} & 0.01 & 2.46  & \textbf{102}\\
    & $4\times 60$  &  104516 & 202338 & \textbf{0.9733} & 0.01 & 4.94  & \textbf{324}\\
    & $4\times 80$  & 140716 & 272818 &  \textbf{0.9733} & 0.01  & 8.45  & \textbf{697}\\
    & $4\times 100$ & 176916 & 343298 &  \textbf{0.9733} & 0.02  & 12.10 & \textbf{1332}\\
    \bottomrule
  \end{tabular}
  }
  \vspace{-1mm}
\end{table}

\begin{table}[t]
 \centering
 \caption{\normalfont Results for \ref{sc:tworooms}}
   \label{tab:pg_rooms}
   \begin{tabular}{crrrrrrrr}
    \toprule
        & \multicolumn{2}{c}{PG} & \multicolumn{3}{c}{Run times} & \multicolumn{1}{c}{MDP} \\
    \#C & States & Choices & Result & Create & Model & Solve & Result \\
    \midrule
    none &   76676 &  145762 & 0.3997 & 0.22 &   9.2 &   \textbf{33.1} & \textbf{0.9999}\\
    2    &  149056 &  292931 & 0.8820 & 0.24 &  23.0 &   \textbf{64.1} & \textbf{0.9999}\\
    \bottomrule
  \end{tabular}
  \vspace{-4mm}
\end{table}

\subsection{Evaluation}
\label{ssec:evaluation}

\noindent Consider \ref{sc:grid}: \prismpomdp{} delivers
results within reasonable time only for very small examples (cf.\ Table~\ref{tab:pomdp_vs_pg}); already for the $6\times 6$ grid it runs out of memory. On the other hand,
our abstraction handles grids up to $30\times30$ within minutes, while still providing strategies with a solid accuracy.
The safety probability is lower for small grids, as there is less room for $\robot$ to
avoid $\cleaner$, and there are proportionally more situations
in which $\robot$ is trapped in a corner or against a wall.

For grids of any size, the safety probability computed on our abstraction is considerably
higher than the na\"\i{}ve approach of ignoring the $\cleaner$ and hoping for the best:
Ignoring the $\cleaner$ during strategy computation yields a probability of just $0.2429$ in the $3\times3$ and $0.8209$ in the $10\times10$ grid,
compared to $0.8323$ and $0.9999$, respectively.

For the MDP, the state space for an $n\times n$ grid is in $\mathcal{O}(n^4)$ compared to a state
space in $\mathcal{O}(r^2n^2)$ for the PG, where $r$ is the viewing range.
As a consequence, no upper bound could be computed for the $50\times 50$ grid, as constructing
the state space yielded a mem-out.

We also compare our results to \solvepomdp, which confirms the results of \prismpomdp for the $3\times 3$ grid,
taking 68~seconds and 107~iterations, but timed out for the $4\times 4$ grid after just 9~iterations.

In Table~\ref{tab:pg_corridor_cross}, for the \ref{sc:corridor} benchmarks, we see that the safety probability
goes down for grids with a longer corridor. The reason for this effect is that, in the
abstraction, $\robot$ can meet $\cleaner$ multiple times when
traveling down the corridor. To avoid this unrealistic behavior, we
applied a simple history refinement as described in Sect.~\ref{ssec:refinement}: we divided the workspace into
four evenly sized regions and have $\robot$ know which region $\cleaner$ was visible in during the last turn.
Thus, $\robot$ ``remembers'' if $\cleaner$ was last seen in front or behind itself.
Since the corridor is narrow enough to always observe its entire width, $\cleaner$, once overtaken,
can never appear in front of $\robot$ again, making the safety probabilities constant independent
of the length of the corridor.
The same effect can be achieved by mapping the strategies obtained in the abstraction back
to the original POMDP: In the resulting MC, the unrealistic behavior is no longer an issue, and safety
probabilities actually increase with the length of the corridor, from $0.9962$
in the $4\times 40$ to $0.9976$ in the $4\times 100$ corridor.

Table~\ref{tab:pg_corridor_cross}, \ref{sc:cross}, indicates that the precomputation of the
visibility-lookup (see Sect.~\ref{sec:case_study}) for large grids with many obstacles eventually
takes significant time, yet the model construction time increases at a faster pace.
In comparison with \ref{sc:grid}, we see that adding obstacles decreases the number of reachable
states and thus also reduces the number of choices and transitions.
Eventually, model construction takes longer than the actual model checking.

Figures~\ref{fig:pg_random_1} and \ref{fig:pg_random_2} show that obstacles --~depending on their number and position~-- can
have varied effects on the results and run times of model checking, although the accumulated results
in Figure~\ref{fig:pg_random_4} show that, in general, model checking times go down as the number of
obstacles increases (and therefore, the total number of reachable states decreases). Safety, on the other hand and as
seen in Figure~\ref{fig:pg_random_3},
is negatively affected as adding more obstacles induces additional blind spots in which $\robot$ can
no longer observe $\cleaner$'s movement. Yet, as witnessed by Set 5, it may also provide safe areas.

Similar blind spots occur in the two rooms example, with results depicted in Table~\ref{tab:pg_rooms} (\ref{sc:tworooms}). 
We add cameras to aid $\robot$ by providing improved visibility around the blind spot, resulting in a near-perfect safety probability. 
The improved visibility doubles the state space size and increases the model checking time by about 40~seconds.

\begin{figure*}[ht]
   \centering
      \subfloat[]{\includegraphics{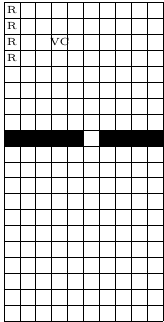}\label{fig:path:step1}}\qquad
      \subfloat[]{\includegraphics{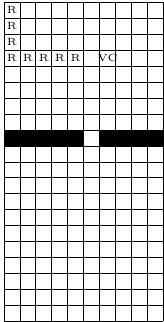}\label{fig:path:step2}}\qquad
      \subfloat[]{\includegraphics{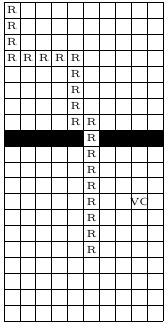}\label{fig:path:step3}}\qquad
      \subfloat[]{\includegraphics{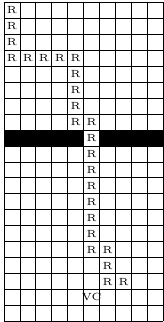}\label{fig:path:step4}}\qquad
      \subfloat[]{\includegraphics{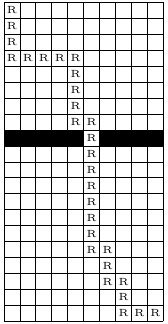}\label{fig:path:step5}}\qquad
   \caption[]{Possible movements of $\robot$ and $\cleaner$ under the strategy computed (for \ref{sc:tworooms}).}\label{fig:path}
\end{figure*}

Finally, we use the two rooms example to visualize the strategy computed by our approach in Fig.~\ref{fig:path}. 
In each subfigure, one can see the steps taken by
$\robot$ so far, as well as the position of $\cleaner$ in the current step as simulated by the abstraction. 
One would expect $\robot$ to try and move away from $\cleaner$ as
far as possible, as seen in Fig.~\ref{fig:path:step2}, but Figs.~\ref{fig:path:step1}, \ref{fig:path:step3} and \ref{fig:path:step4} actually show the opposite:
$\cleaner$ appears on the edge of $\robot$'s view range and $\robot$ moves towards it, as, due to the abstraction, $\cleaner$'s behavior
gets actually more powerful when it cannot be observed. In the abstraction, it is
beneficial for $\robot$ to keep $\cleaner$ within the view range, where the $\cleaner$ behavior is merely probabilistic and 
not non-deterministic.

\section{Discussion}
\label{sec:discussion}

\noindent Game-based abstraction successfully prunes the state space of MDPs by
merging similar states. By adding an adversary that assumes the
worst-case state, a PG is obtained. In general, abstraction turns the POMDP at hand into a
partially observable PG, which remains intractable. However, splitting according to
observational equivalence leads to a fully observable PG. PGs can be analyzed by
black-box algorithms as implemented, \eg in \prismgames, which also returns
an optimal strategy. The strategy from the PG can be applied to the POMDP, which
yields the actual (higher) safety level.

In general, the abstraction can be too coarse; however, in the examples above, we have successfully shown
that the game-based abstraction is not too coarse if one makes some assumptions
about the POMDP. These assumptions are often naturally fulfilled by motion planning scenarios.

The assumptions from Sect.~\ref{ssec:problem_statement} can be relaxed in several
respects: Our method naturally extends to \emph{multiple moving obstacles}. We restricted the method
to a single controllable agent, but if information is shared among multiple agents, the
method is applicable also to this setting. If information sharing is
restricted, special care has to be taken to prevent information leakage. Richer
classes of behavior for the obstacles, including non-deterministic
choices, are an important area for future research. Non-deterministic moving obstacles, for instance, lead to partially observable
PGs, and game-based abstraction yields three-player games. As two sources of
non-determinism are uncontrollable, both the obstacles and the abstraction can be
controlled by Player~2, thus yielding a PG again.

Supporting a richer class of specifications is another option: \prismgames{}
supports a probabilistic variant of alternating (linear-) time logic extended by rewards
and trade-off analysis. With the same abstraction technique presented here, a
larger class of properties can be analyzed. However, care has to be taken when
combining invariants and reachability criteria arbitrarily, as they involve under- and
over-approximations.

Our method can be generalized to POMDPs for other settings. We use the original
problem statement on the graph only to motivate the correctness.
The abstraction can be lifted (as indicated by Def.~\ref{def:abstract_pg}),
for refinement, however, a more refined argument for correctness is necessary.

The proposed PG construction is straightforward and currently realized without constructing the POMDP first.
This approach simplifies the implementation of the refinement, as for large grids the POMDP can be considerably
larger than the abstraction. If the POMDP is small enough to handle, we can build it, too, and map the strategy
from the abstraction to the original model to obtain the precise safety probability (cf.\ the $(p+\tau)$-safety in Fig.~\ref{fig:scheme}).

\section{Conclusion}

\noindent We developed a game-based abstraction technique to synthesize strategies for a class of POMDPs.
This class encompasses typical grid-based motion planning problems under restricted observability of the environment.
For these scenarios, we efficiently compute strategies that allow the agent to maneuver the grid in order to reach a given
goal state while at the same time avoiding collisions with faster moving obstacles.
Experiments show that our approach can handle state spaces up to three orders of magnitude larger than general-purpose state-of-the-art POMDP solvers in less time,
while at the same time using less states to represent the same grid sizes.

\balance
\bibliographystyle{IEEEtran}
\bibliography{abbrev_short,literature}

\clearpage
\nobalance









\end{document}